\documentclass[10pt,twocolumn,letterpaper]{article}
\usepackage{lineno}
\usepackage[pagenumbers]{cvpr}
\usepackage{graphicx}
\usepackage{amsmath}
\usepackage{amssymb}
\usepackage{booktabs}
\usepackage{blindtext}

\def\paperID{*****} 
\def\confName{CVPR}
\def\confYear{2025}

\title{Deep neural network-based detection \\ of counterfeit products from smartphone images}

\author{Hugo Garcia-Cotte, Dorra Mellouli,
Abdul Rehman, and
Li Wang
\\ 
Cypheme
\\27 rue Bargue 75015 Paris, France\\
{\tt\small hugo@cypheme.com}
\and
David G.~Stork\\
Adjunct Professor\\
Stanford University\\
{\tt\small dstork@stanford.edu}
}

\begin{document}
\maketitle
\begin{abstract}
 Counterfeit products such as drugs and vaccines as well as  luxury items such as high-fashion handbags, watches, jewelry, garments, and cosmetics, represent significant direct losses of revenue to legitimate manufacturers and vendors, as well as indirect costs to societies at large. We present the world\rq s first purely computer-vision-based system to combat such
 counterfeiting---one that does not require special security tags or other alterations to the products or modifications to supply chain tracking.  Our deep neural network system shows high accuracy on branded garments from our first manufacturer tested (99.71\% after 3.06\% rejections) using images captured under natural, weakly controlled conditions, such as in retail stores, customs checkpoints, warehouses, and outdoors. Our system, suitably transfer trained on a small number of fake and genuine articles, should find application in additional product categories as well, for example fashion accessories, perfume boxes, medicines, and more.
\end{abstract}    
\section{The problem of counterfeit products}
\label{sec:intro}

 Counterfeit drugs, medical systems, and their associated effects are responsible for 1M human deaths worldwide annually, with over 500,000 deaths in sub-Sahara Africa alone, according to the United Nations Office on Drugs and Crime.\cite{Fleck:23}  Counterfeit products—particularly luxury items such as high-fashion handbags, watches, jewelry, garments, as well as drugs and vaccines—represent a direct loss of income to legitimate manufacturers and retailers of over \$0.5T/year worldwide; fake merchandise online in certain commercial sectors, such a clothing, is growing at nearly 50\% per year.\cite{kennedy2020counterfeit}  A recent study involving 13,000 consumers from 17 countries found that 74\% of people have purchased counterfeit products in the previous year.\cite{Alhabashetal:23}
 
 The true costs of counterfeits are much higher than direct loss to legitimate manufacturers and retailers, however, bringing the full costs to an estimated \$1.8T. For instance, revenues from counterfeit sales often fund other illegal enterprises, including human trafficking, drug dealing, and so on, each with their associated financial, human, and societal costs. Most counterfeit products fail safety regulations and other compliance requirements leading to significant indirect costs.  For instance, lithium-ion batteries (one of the most counterfeited products, worldwide) are far more prone to explosion, fire, leakage, and concomitant damage to devices such as personal computers, mobile phones, and appliances than are legitimate batteries.\cite{WinNT}  A pallet of counterfeit lithium-ion batteries burst into flames on a cargo flight in 2010 and the plane then crashed, killing all the crew.\cite{Hradecky:10}

 This is not the place to review the broad range of anti-counterfeiting technologies, applicable to the wide array of products, currency bills, valued documents (wills, diplomas), medical packages, and so on.\cite{yang2017effectiveness} Suffice it to say, however, that the majority of anti-counterfeiting approaches involve special tags, stamps, or alterations to products themselves.  Other, related, methods employ marks that are read with special detectors or illumination such as ultraviolet light. While
 these methods can indeed lead to high accuracy, they often involve additional production and monitoring costs, cannot apply to previously manufactured goods, and have other drawbacks.\cite{zhu2014design}  Some manufacturers, vendors, and law enforcement organizations rely on human experts who examine potential counterfeit goods.  This approach yields inconsistent accuracy, high cost, and of course is not scalable.
 
 Our application goals, and hence our overall system, overcome many of the drawbacks in such prior approaches in large part because of the recent  progress in deep neural network visual pattern classification.\cite{goodfellow2016deep}  In brief, we transfer train a deep neural network image classifier with a relatively small corpus of images of genuine and of counterfeit products.  In Sects.~\ref{sec:Goals},~\ref{sec:System}, and \ref{sec:Results}, we shall illustrate our approach using brand marks such as emblems on polo shirt logos, but our approach is applicable to much greater range of counterfeit products.

An important aspect of our research is to explore a range of deep neural net architectures, with different number of layers, neurons, and structures, to find one that is most appropriate for our application.  Table~\ref{tbl:Results}, below, summarizes our results on this aspect of our research.
\section{Application goals and system specifications}
\label{sec:Goals}
Our goal is to create an anti-counterfeiting system for products based on high-accuracy classification of products (as {\sc counterfeit} or {\sc genuine}) having the following properties:

\begin{itemize}
    \item it requires no special tags or other attachments to the product (as these can often be removed, add costs and delays to manufacture, and cannot be applied to previously released products)
    \item requires no alteration to the product itself (as this adds cost and delays to manufacture and may compromise overall design and brand recognition)
    \item works for previously manufactured products already in the marketplace 
    \item does not require involvement from the manufacturer (so can be used by online retailers, governments, safety and compliance regulators, rapidly and without legal or other contractual requirements from the manufacturer)
    \item relies entirely on visual information that can be captured with an internet-connected smartphone camera (as this permits inexpensive testing at nearly every stage of the life of the product and incurs no additional cost for special detection hardware)
    \item is scalable to millions of authentication tests per day (to keep up with the rapid expansion of counterfeit products)
    \item is robust to adversarial counter-measures
    \item is highly accurate, that is has small type~I and type~II errors and can be tailored to different cost matrices (different costs for type~I and type~II errors)
    \item can be updated and improved in response to the \lq\lq arms race\rq\rq\ of counterfeiters developing ever-more difficult-to-detect counterfeit products
\end{itemize}
 
Our deep neural network system satisfies all these design desiderata, as we shall see.
\section{Recognition system, image data, and
 training and testing protocols} \label{sec:System}

\begin{figure} 
\begin{center}
\includegraphics[width=.35\textwidth]{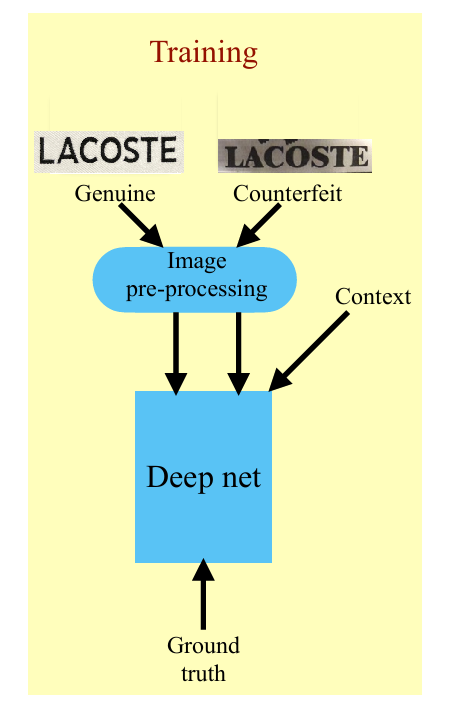}
\end{center}
\caption{\label{fig:Flowchart}Flowchart of our anti-counterfeiting system.  A base image-classification deep network is transfer trained with a relatively small ($\approx 20,000$) examples of ground truth images of genuine and counterfeit product marks, preprocessed for affine transforms such as rotation, translations, and scale.  The trained net is used to classify a query item as {\sc genuine} or {\sc counterfeit}.}
\end{figure}
 Figure~\ref{fig:Flowchart} shows the system architecture and flowchart of our anti-counterfeiting system.  During training natural smartphone images of brand marks were captured, and processed by simple affine transformations to consistent overall scale, orientation, and alignment of processed images.

Our approach is to use deep neural network classification applied to smartphone images, as this fulfills several of our design specifications. Images are captured by smartphones followed by mark or emblem localization, alignment, and classification.  For localization we did a transfer learning on efficientdet D0 model to be able to detect the mark (here, crocodile) from the neck tag on the product.  For alignment we used a Spatial Transformer Network (STN) which is a learnable module that can be placed in a Convolutional Neural Network (CNN), to increase the spatial invariance in an efficient manner.  Spatial invariance refers to the invariance of the model towards spatial transformations of images such as rotation, translation and scaling not fully corrected by our preprocessing steps.  Invariance is the ability of the model to recognize and identify features even when the input is transformed or slightly modified.

Our general system can take as input non-visual contextual information about the product, the location, capture venue (street vendor, commercial shop, customs inspection station, etc.) and learned priors describing probabilities of counterfeits of a particular product or product class.  We report here the performance of the system without such contextual information, that is, based solely on the visual information captured from the product.
 
\begin{figure}
\begin{center}
\begin{tabular}{cc}
\includegraphics[width=.2\textwidth]{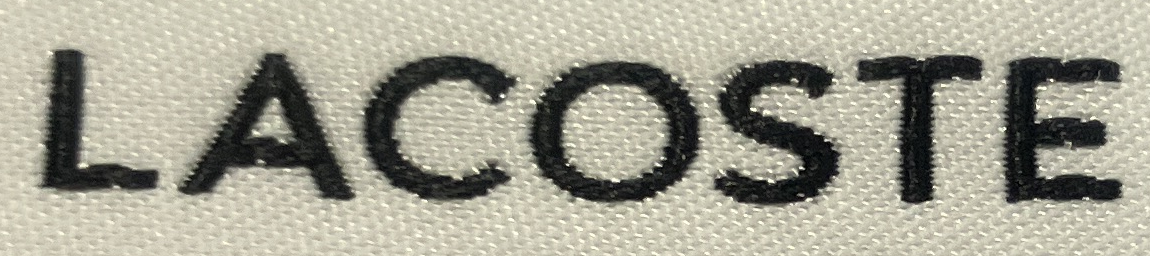} &
\includegraphics[width=.2\textwidth]{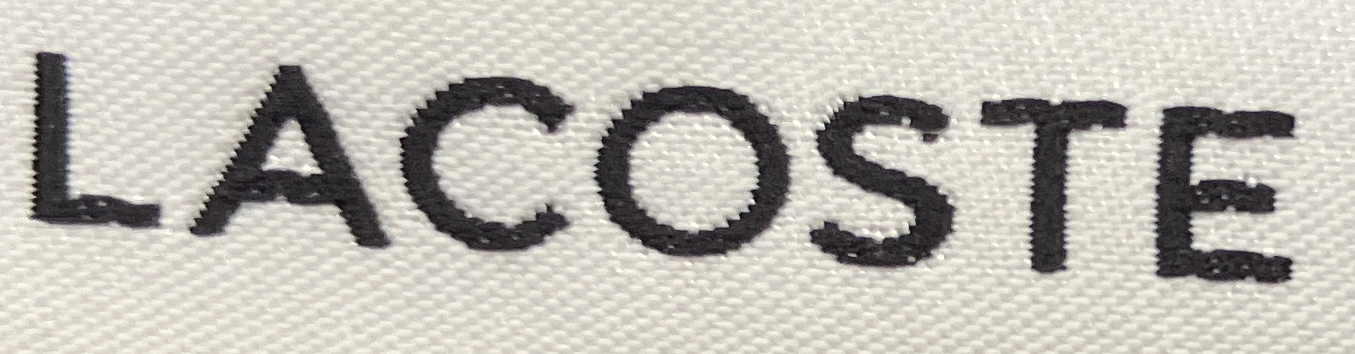} \\
\includegraphics[width=.2\textwidth]{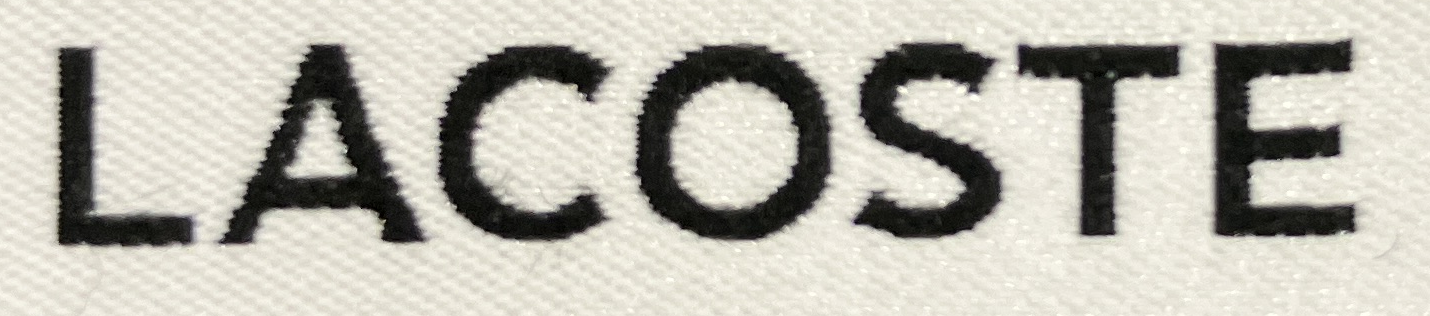} &
\includegraphics[width=.2\textwidth]{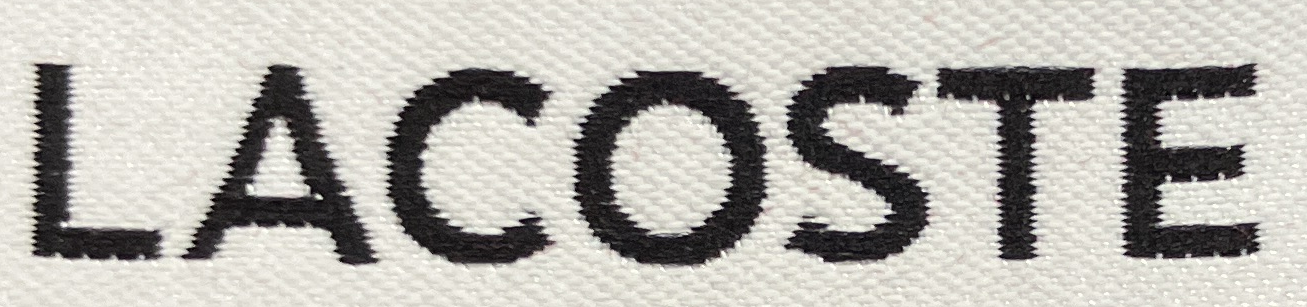} \\
\hline
\includegraphics[width=.2\textwidth]{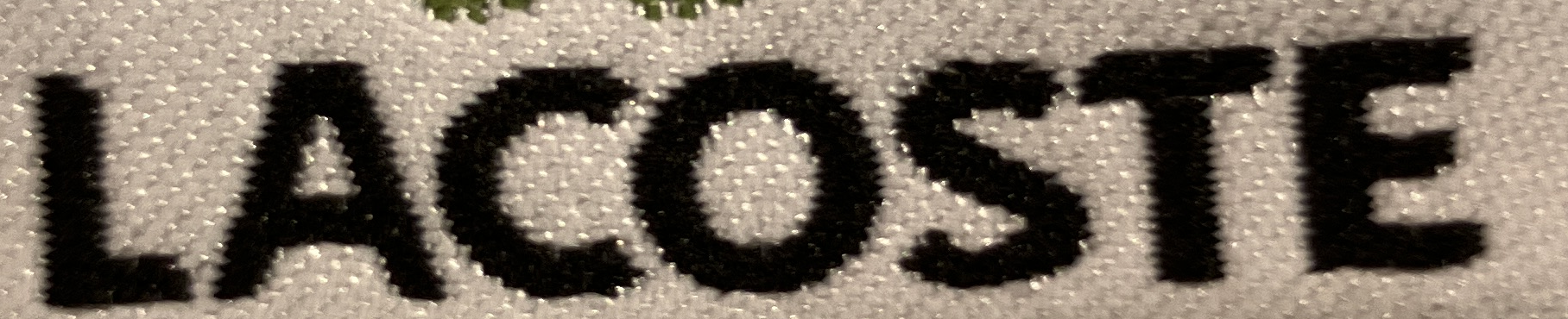} &
\includegraphics[width=.2\textwidth]{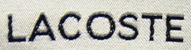} \\
\includegraphics[width=.2\textwidth]{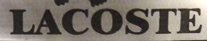} &
\includegraphics[width=.2\textwidth]{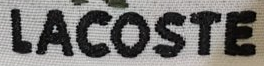} \\
\end{tabular}
\end{center}
\caption{\label{fig:Lacoste}(T)~Images of genuine polo shirt emblems and (B)~of counterfeit emblems.  }
 \end{figure}

 Figure~\ref{fig:Lacoste} shows smartphone images of logos from our
 training database taken by non-expert users: four genuine Lacoste polo shirts and four fakes. Ground truth for fakes was provided by human experts from garment manufacturer Lacoste.  Additional classification was performed by our cloud-based system and was so fast it did not slow Lacoste\rq s operations.
 Processing is done in the cloud, and thus is fully scalable to millions of queries, or more, per day from any location with cloud access.

  We explored nine architectures of deep networks, chosen to ensure a variety of network architectures and sizes, and to explore if integration of multiple classification results would lead to improved
 overall accuracy:
 \begin{itemize}
     \item ConvNext-small \cite{yu2024inceptionnext} 
     \item ConvNext-base\cite{yu2024inceptionnext} 
     \item Swim-transformer base\cite{liu2021swin}
     \item Twins-SVT-base \cite{chu2021twins}
     \item Twins-SVT-large\cite{chu2021twins}
     \item Twins-PCPVT-large\cite{chu2021twins}
     \item ViT-S\cite{alexey2020image}
     \item ViT-L\cite{alexey2020image}
     \item AntiCounterfeit
 \end{itemize}

 We used 20,945 images for training, 8747 for validation, and 1702 for testing.\cite{Duda2001pattern}  To train and validate our system, we collaborated with Lacoste to assemble a diverse dataset. Lacoste provided examples of genuine, fake, and \lq\lq super fake\rq\rq\ Lacoste polos. Our team then captured images of these items using various smartphones in different, uncontrolled environments, including diverse lighting conditions and angles, to simulate the varied scenarios in which counterfeit detection may occur. This allowed us to create a dataset that closely mimics real-world conditions.

Additionally, we obtained images from customs officers around the world, who regularly send photos to Lacoste for authentication. These images provided an additional layer of diversity to the dataset, representing the wide range of circumstances under which counterfeits are encountered globally.

For final testing, we evaluated the system in a live operational setting at Lacoste\rq s processing facility in Troyes, France, where online returns of items purchased through Lacoste\rq s e-commerce platform are handled. As part of the repackaging process, staff at this facility used a web application to authenticate returned polos.
 
 All images---for training and testing---were captured under weakly controlled conditions with a variety of mobile phone cameras.  This approach matched the weakly controlled conditions obtained in a fielded system and thereby reduced bias and improved accuracy.  We used binary cross-entropy as a loss function and early stopping to avoid overfitting.  We trained and tested our system on NVIDIA A100 GPUs through Google Colab Pro+, developed using the PyTorch library. 
 
\section{Results} \label{sec:Results}

We found that the most-accurate classification system was a convolutional network with 380 layers and 133M weights total, which achieved 99.71\% accuracy after rejecting 3.06\% of patterns as ambiguous.  This high classification rate was confirmed by the product manufacturer.\footnote{One of our classification deep networks, including weights, and samples of real and fake images will be available as open source upon request for non-commercial experimentation and analysis.}  

\begin{table*}[ht] 
\begin{center}
\begin{tabular}{|r|cccc|}
\hline
architecture & layers & weights & rejection & accuracy \\ 
\hline
ConvNext-small & 50 & 50M & 3.54\% & 98.16\% \\
ConvNext-base & 50 & 89M & 0\% & 98.95\% \\
Swim-transformer base & 12 transformer blocks
& 88M & 8.13\% & 97.70\% \\
Twins-SVT-base & 12 transformer blocks & 56M & 10.46\% & 96.94\% \\
Twins-SVT-large & 12 transformer blocks & 99M & 12.15\% & 90.01\% \\
Twins-PCPVT-large & 12 transformer blocks & 76M & 4.34\% & 98.65\% \\
ViT-S & 12 transformer blocks & 22M & 3.54\% & 95.91\% \\
ViT-L & 24 transformer blocks & 307M & 3.30\% & 98.58\% \\
AntiCounterfeit & 380 & 133M & 3.06\% & 99.71\%\\
\hline
\end{tabular}
\end{center}
\caption{\label{tbl:Results}The deep net architecture, the number of layers, total number of connection weights, and rejection rate, overall classification accuracy on a genuine/fake recognition task based on mobile phone images of 1702 polo shirt emblems.}
\end{table*}

Our AntiCounterfeit convolutional network performed slightly better---in small rejection rate and high classification accuracy---as the {ViT-L} transformer network, with roughly $1/3$ the number of connection weights.  

The output of our system is a scalar in the range $0 \leftrightarrow 1$ for any test image.  In the most direct classification rule, we simply classify a pattern as {\sc genuine} if that output is greater that $0.5$, and {\sc counterfeit} otherwise.  However, users can adjust these decisions thresholds given a cost matrix, which expresses the overall costs due to type~1 and type~2 errors, or the cost of rejecting an image due to insufficient or ambiguous visual evidence. 

\begin{figure}
\includegraphics[width=0.45\textwidth]{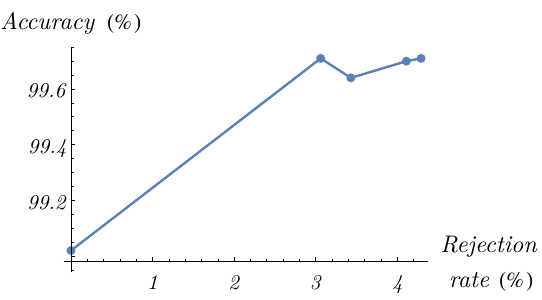}
\caption{\label{fig:Tradeoff}The empirical tradeoff between rejection rate and classification accuracy for our AntiCounterfeit deep network classifier.}
\end{figure}

Figure~\ref{fig:Tradeoff} shows the effect of adjusting such thresholds in our AntiCounterfeit system, specifically the dependency of the overall accuracy as a function of rejection rate.  We found that the highest accuracy was 99.71\% at a 3.06\% rejection rate for this task.  This result gave the largest total number of correctly classified patterns.  The specific result is highly unlikely to generalize---that is, the empirical accuracy versus rejection rate will be task specific.

A small percentage of counterfeit products are so-called \lq\lq super fakes,\rq\rq\ typically made with extreme care;  these can sometimes escape even expert human evaluators intimately familiar with the product at hand.  For this reason our system promises to be more accurate than even experts, of course with the numerous other benefits listed in Sect.~\ref{sec:Goals}.
\section{Conclusion} \label{sec:Conclusions}

We have demonstrated a high-accuracy (99.71\%) anti-counterfeiting system based on deep neural network image recognition.  Apparently, deep net architecture learns to distinguish between the \lq\lq noise\rq\rq\ variations in an image of a logo due to lighting, manufacturing variability of color, stitching, shape, with \lq\lq signal\rq\rq\ variations due to counterfeit manufacture.  We can further train our fielded system with newly classified images and thus improve accuracy in the \lq\lq arms race\rq\rq\ against counterfeiters.  Although our results reported here did not include contextual information or such retraining, this information has proven useful in related classification problems and would likely improve our results.

A very large proportion of classes of counterfeit products share the same underlying structure (variations due to legitimate manufacture and variations due to counterfeiting) and thus we are confident our approach is applicable to a very wide range of counterfeit products, with associated savings in direct and indirect costs to producers, vendors, and societies at large.  

Finally, while current anti-counterfeitting systems rely heavily on a small number of human experts, typically within the offices of a manufacturer, our technology promises to empower millions of end users with our app, worldwide, to act as super authenticators.

\section*{Acknowledgements}  This project was funded by Horizon 2020.  We thank our partner, Lacoste S.A., Paris, France for image data, human classification results, and insights.
{
    \small
    \bibliographystyle{ieeenat_fullname}
    \bibliography{main.bib}
}
\clearpage
\maketitleagain{Supplementary Material}

The supplementary can be compiled together with the main paper, therefore:
\begin{itemize}
\item The supplementary can back-reference sections of the main paper, for example, we can refer to \cref{sec:intro};
\item The main paper can forward reference sub-sections within the supplementary explicitly (e.g. referring to a particular experiment); 
\item When submitted to arXiv, the supplementary will already included at the end of the paper.
\item To submit the supplementary, use the \verb|\includeonly{| command to avoid typesetting the paper itself.
\item You can also add citations that you may have missed~\cite{Alpher2005}.
\end{itemize}

{
    \small
    \bibliographystyle{ieeenat_fullname}
    \bibliography{main.bib}
}

\clearpage
\maketitleagain{Guidelines for Author Response}
\setcounter{page}{1}
\setcounter{section}{1}

After receiving paper reviews, authors may optionally submit a rebuttal to address the reviewers' comments, which will be limited to a {\bf one page} PDF file.
Please follow the steps and style guidelines outlined below.

The author rebuttal is optional and, following similar guidelines to previous conferences, is meant to provide you with an opportunity to rebut factual errors or to supply additional information requested by the reviewers.
It is NOT intended to add new contributions (theorems, algorithms, experiments) that were absent in the original submission and NOT specifically requested by the reviewers.
You may optionally add a figure, graph, or proof to your rebuttal to better illustrate your answer to the reviewers' comments.

Per a passed 2018 PAMI-TC motion, reviewers should refrain from requesting significant additional experiments for the rebuttal or penalize for lack of additional experiments.
Authors should refrain from including new experimental results in the rebuttal, especially when not specifically requested to do so by the reviewers.
Authors may include figures with illustrations or comparison tables of results reported in the submission/supplemental material or in other papers.

Just like the original submission, the rebuttal must maintain anonymity and cannot include external links that reveal the author identity or circumvent the length restriction.
The rebuttal must comply with this template (the use of sections is not required, though it is recommended to structure the rebuttal for ease of reading).


\paragraph{Response length}
Author responses must be no longer than 1 page in length including any references and figures.
Overlength responses will simply not be reviewed.
This includes responses where the margins and formatting are deemed to have been significantly altered from those laid down by this style guide.
Note that this \LaTeX\ guide already sets figure captions and references in a smaller font.

\section{Formatting your response}
{\bf Make sure to update the paper title and paper ID in the appropriate place in the tex file.}
All text must be in a two-column format.
The total allowable size of the text area is $6\frac78$ inches (17.46 cm) wide by $8\frac78$ inches (22.54 cm) high.
Columns are to be $3\frac14$ inches (8.25 cm) wide, with a $\frac{5}{16}$ inch (0.8 cm) space between them.
The top margin should begin 1 inch (2.54 cm) from the top edge of the page.
The bottom margin should be $1\frac{1}{8}$ inches (2.86 cm) from the bottom edge of the page for $8.5 \times 11$-inch paper;
for A4 paper, approximately $1\frac{5}{8}$ inches (4.13 cm) from the bottom edge of the page.

Please number any displayed equations.
It is important for readers to be able to refer to any particular equation.

Wherever Times is specified, Times Roman may also be used.
Main text should be in 10-point Times, single-spaced.
Section headings should be in 10 or 12 point Times.
All paragraphs should be indented 1 pica (approx.~$\frac{1}{6}$ inch or 0.422 cm).
Figure and table captions should be 9-point Roman type as in \cref{fig:rebuttal}.

List and number all bibliographical references in 9-point Times, single-spaced, at the end of your response.
When referenced in the text, enclose the citation number in square brackets, for example~\cite{Alpher05}.
Where appropriate, include the name(s) of editors of referenced books.

\begin{figure}[t]
  \centering
  \fbox{\rule{0pt}{0.5in} \rule{0.9\linewidth}{0pt}}
   \caption{Example of caption.  It is set in Roman so that mathematics
   (always set in Roman: $B \sin A = A \sin B$) may be included without an
   ugly clash.}
   \label{fig:rebuttal}
\end{figure}

To avoid ambiguities, it is best if the numbering for equations, figures, tables, and references in the author response does not overlap with that in the main paper (the reviewer may wonder if you talk about \cref{fig:rebuttal} in the author response or in the paper).
See \LaTeX\ template for a workaround.

\subsection{Illustrations, graphs, and photographs}

All graphics should be centered.
Please ensure that any point you wish to make is resolvable in a printed copy of the response.
Resize fonts in figures to match the font in the body text, and choose line widths which render effectively in print.
Readers (and reviewers), even of an electronic copy, may choose to print your response in order to read it.
You cannot insist that they do otherwise, and therefore must not assume that they can zoom in to see tiny details.

When placing figures in \LaTeX, it is almost always best to use \verb+\includegraphics+, and to specify the  figure width as a multiple of the line width as in the example below
{\small\begin{verbatim}
   \usepackage{graphicx} ...
   \includegraphics[width=0.8\linewidth]
                   {myfile.pdf}
\end{verbatim}
}

{
    \small
    \bibliographystyle{ieeenat_fullname}
    \bibliography{main.bib}
}
\end{document}